\documentclass[12pt]{article}
\usepackage[top=0.95in,bottom=0.95in,left=0.95in,right=0.95in]{geometry}

\usepackage[utf8]{inputenc} % allow utf-8 input
\usepackage[T1]{fontenc}    % use 8-bit T1 fonts
\usepackage{hyperref}       % hyperlinks
\usepackage{url}            % simple URL typesetting
\usepackage{booktabs}       % professional-quality tables
\usepackage{amsfonts}       % blackboard math symbols
\usepackage{nicefrac}       % compact symbols for 1/2, etc.
\usepackage{microtype}      % microtypography
\usepackage{xcolor}         % colors

\title{Metric Elicitation; Moving from Theory to Practice}

\author{%
  Safinah Ali\thanks{Equal Contribution} \\
  Massachusetts Institute of Technology \\
  \texttt{safinah@media.mit.edu} \\
  \and
  Sohini Upadhyay\samethanks \\
  Harvard University \\
  \texttt{supadhyay@g.harvard.edu} \\
  \and
  Gaurush Hiranandani \\
  Amazon \\
  \texttt{hgaurush@amazon.com} \\
  \and
  Elena L. Glassman\\
 Harvard University\\
    \texttt{glassman@seas.harvard.edu} \\
    \and
    Oluwasanmi Koyejo\\
    Stanford University \& Google \\
    \texttt{sanmi@stanford.edu} 
}

\usepackage{graphicx}
\usepackage{subfigure}
\usepackage{caption}[justification=justified,singlelinecheck=false]
\usepackage{diagbox}
\usepackage{wrapfig}

\usepackage{amsmath}
\usepackage{dsfont}
\usepackage{amssymb,commath}

\usepackage{bm,xfrac,xcolor}
\usepackage{nccmath}
\usepackage[toc,page,header]{appendix}

\usepackage{algorithm}
\usepackage[noend]{algorithmic}
\usepackage{bbm}
\usepackage{enumitem}
\usepackage{refcount}
\usepackage{mathtools}
\usepackage{stackengine}
\usepackage{multirow}
\usepackage{makecell}

\usepackage{comment}
\usepackage{mathrsfs}
\makeatletter
\newcommand*{\addFileDependency}[1]{% argument=file name and extension
  \typeout{(#1)}
  \@addtofilelist{#1}
  \IfFileExists{#1}{}{\typeout{No file #1.}}
}
\makeatother


\newcommand*\samethanks[1][\value{footnote}]{\footnotemark[#1]}

\def\myfnt{\ifx\protect\@typeset@protect\expandafter\footnote\else\expandafter\@gobble\fi}
\makeatother

\makeatletter
\def\BState{\State\hskip-\ALG@thistlm}
\makeatother

\usepackage{placeins,afterpage}

\makeatletter
\newcommand*\bigcdot{\mathpalette\bigcdot@{.5}}
\newcommand*\bigcdot@[2]{\mathbin{\vcenter{\hbox{\scalebox{#2}{$\m@th#1\bullet$}}}}}
\makeatother

\usepackage{tikz}
\tikzset{
  c/.style={every coordinate/.try}
}

\mathchardef\mhyphen="2D

\newcommand{\Mcal}{{\cal M}}

\newcommand{\Xcal}{{\cal X}}

\newcommand{\Pmbb}{\mathbb{P}}

\newcommand{\cmbfbar}{\oline{\mathbf{c}}}

\newcommand{\ambf}{\mathbf{a}}

\newcommand{\cmbf}{\mathbf{c}}

\newcommand{\oline}[1]{\mkern 1.5mu\overline{\mkern-1.5mu#1}}

\renewcommand{\hbar}{\oline{h}}

\newcommand{\1}{{\mathbf 1}}

\newcommand{\baligned}     {\begin{aligned}}
	\newcommand{\ealigned}     {\end{aligned}}
\newcommand{\barray}       {\begin{array}}
	\newcommand{\earray}       {\end{array}}
\newcommand{\bbmatrix}     {\begin{bmatrix}}
	\newcommand{\ebmatrix}     {\end{bmatrix}}
\newcommand{\bcases}       {\begin{cases}}
	\newcommand{\ecases}       {\end{cases}}
\newcommand{\bcenter}      {\begin{center}}
	\newcommand{\ecenter}      {\end{center}}
\newcommand{\bcolumn}      {\begin{column}}
	\newcommand{\ecolumn}      {\end{column}}
\newcommand{\bcolumns}     {\begin{columns}}
	\newcommand{\ecolumns}     {\end{columns}}
\newcommand{\benumerate}   {\begin{enumerate}}
	\newcommand{\eenumerate}   {\end{enumerate}}
\newcommand{\bequation}    {\begin{equation}}
	\newcommand{\eequation}    {\end{equation}}
\newcommand{\bequationn}   {\begin{equation*}}
	\newcommand{\eequationn}   {\end{equation*}}
\newcommand{\bfigure}      {\begin{figure}}
	\newcommand{\efigure}      {\end{figure}}
\newcommand{\bflushright}  {\begin{flushright}}
	\newcommand{\eflushright}  {\end{flushright}}
\newcommand{\bitemize}     {\begin{itemize}}
	\newcommand{\eitemize}     {\end{itemize}}
\newcommand{\bpmatrix}     {\begin{pmatrix}}
	\newcommand{\epmatrix}     {\end{pmatrix}}
\newcommand{\bsubequations}{\begin{subequations}}
	\newcommand{\esubequations}{\end{subequations}}
\newcommand{\btable}       {\begin{table}}
	\newcommand{\etable}       {\end{table}}
\newcommand{\btabular}     {\begin{tabular}}
	\newcommand{\etabular}     {\end{tabular}}
\newcommand{\bvmatrix}     {\begin{vmatrix}}
	\newcommand{\evmatrix}     {\end{vmatrix}}

\newcommand{\bequali}      {\bsubequations\begin{align}}
	\newcommand{\eequali}      {\end{align}\esubequations}

\newcommand{\balgorithm}  {\begin{algorithm}}
	\newcommand{\ealgorithm}  {\end{algorithm}}
\newcommand{\balgorithmic}{\begin{algorithmic}}
	\newcommand{\ealgorithmic}{\end{algorithmic}}
\newcommand{\bassumption} {\begin{assumption}}
	\newcommand{\eassumption} {\end{assumption}}
\newcommand{\bcorollary}  {\begin{corollary}}
	\newcommand{\ecorollary}  {\end{corollary}}
\newcommand{\bdefinition} {\begin{definition}}
	\newcommand{\edefinition} {\end{definition}}
\newcommand{\bexample}    {\begin{example}}
	\newcommand{\eexample}    {\end{example}}
\newcommand{\bprop}    {\begin{prop}}
	\newcommand{\eprop}    {\end{prop}}
\newcommand{\blemma}      {\begin{lemma}}
	\newcommand{\elemma}      {\end{lemma}}
\newcommand{\bproblem}    {\begin{problem}}
	\newcommand{\eproblem}    {\end{problem}}
\newcommand{\bproof}      {\begin{proof}}
	\newcommand{\eproof}      {\end{proof}}
\newcommand{\bremark}     {\begin{remark}}
	\newcommand{\eremark}     {\end{remark}}
\newcommand{\btheorem}    {\begin{theorem}}
	\newcommand{\etheorem}    {\end{theorem}}

\usepackage[customcolors]{hf-tikz}
\usepackage{adjustbox}
\usepackage[normalem]{ulem}
\usepackage{cancel}
\usetikzlibrary{calc,shapes.geometric}

\begin{document}

\maketitle

\begin{abstract}
Metric Elicitation (ME) is a framework for eliciting classification metrics that better align with implicit user preferences based on the task and context. The existing ME strategy so far is based on the assumption that users can most easily provide preference feedback over classifier statistics such as confusion matrices. This work examines ME, by providing a first ever implementation of the ME strategy. Specifically, we create a web-based ME interface and conduct a user study that elicits users' preferred metrics in a binary classification setting. We discuss the study findings and present guidelines for future research in this direction. 
\end{abstract}
\section{Introduction}
\label{sec:intro}

% \vskip -0.4cm
\emph{Given a classification task, which performance metric should the classifier optimize?} This question is often faced by practitioners while developing machine learning solutions. For example, consider cancer diagnosis where a doctor applies a cost-sensitive predictive model to classify patients into cancer categories~\cite{yang2014multiclass}. 
Although it is clear that the chosen costs directly determine the model decisions, it is not clear how to quantify the expert's intuition into precise quantitative cost trade-offs, i.e., the performance metric~\cite{Dmitriev2016MeasuringM, zhang2020joint}.

% \begin{wrapfigure}{r}{0.6\textwidth}
% \vspace{-0.425cm}
%   \centering
%   \includegraphics[width=0.6\columnwidth]{plots/ME.png}
%   \vspace{-0.3cm}
%   \caption{Metric Elicitation framework~\cite{hiranandani2018eliciting}.}
%   \label{fig:meframework}
%   \vspace{-0.15cm}
% \end{wrapfigure}

\begin{figure}[t]
% {r}{0.6\textwidth}
% \vspace{-0.425cm}
  \centering
  \includegraphics[scale=0.6]{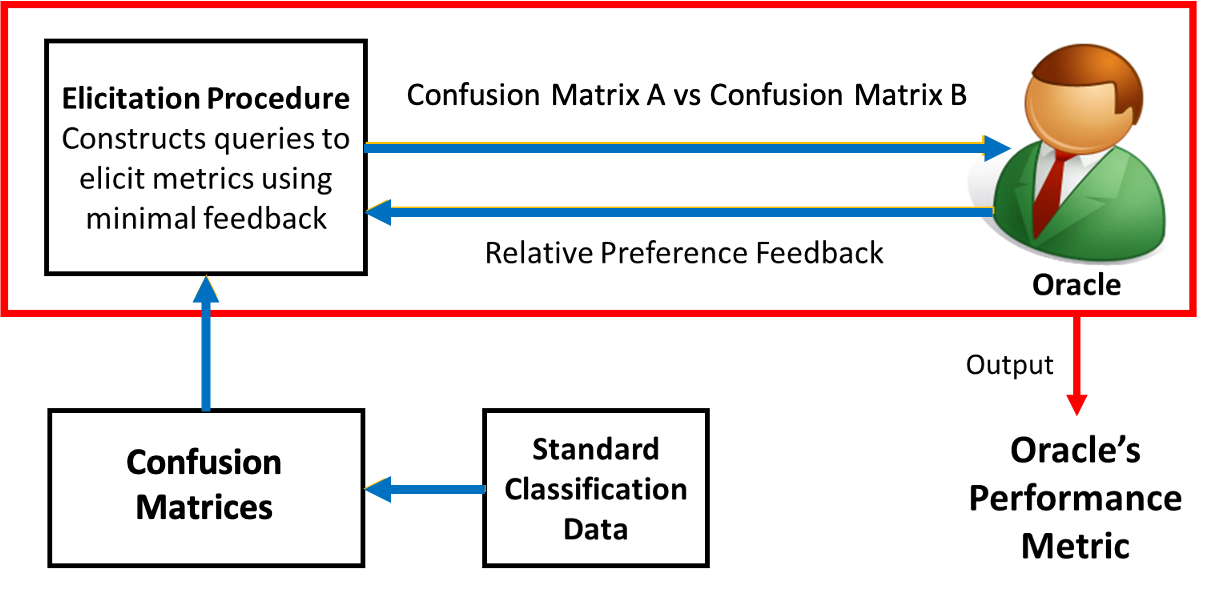}
  \vspace{-0.1cm}
  \caption{Metric Elicitation framework~\cite{hiranandani2018eliciting}.}
  \label{fig:meframework}
  \vspace{-0.15cm}
\end{figure}

Hiranandani et al.~\cite{hiranandani2018eliciting, hiranandani2019multiclass} addressed this issue by formalizing the \emph{Metric Elicitation (ME)} framework, whose goal is to estimate a performance metric using user feedback over confusion matrices.
% from a user. 
The motivation is that by employing metrics that reflect a user's innate trade-offs given the task, context, and population at hand, one can learn models that best capture the user preferences~\cite{hiranandani2018eliciting}. As humans are often inaccurate in providing absolute quality feedback~\cite{qian2013active}, Hiranandani et al.~\cite{hiranandani2018eliciting} propose to use pairwise comparison queries, where the user (oracle) is asked to provide a relative preference over two confusion matrices. 
% The motivation is that by employing metrics that reflect a user's innate trade-offs given the task, context, and population at hand, one can learn models that best capture the user preferences~\cite{hiranandani2018eliciting}. 
% As humans are often inaccurate in providing absolute quality feedback~\cite{qian2013active}, Hiranandani et al.~\cite{hiranandani2018eliciting} propose to use pairwise comparison queries, where 
% In ME, the user (oracle) is asked to compare two confusion matrices and provide a relative preference. Using such pairwise comparison queries, ME aims to recover the oracle's metric. 
Figure~\ref{fig:meframework} (reproduced from~\cite{hiranandani2018eliciting})  depicts the ME framework. 

% \marginpar{%
%   \vspace{-10cm} \fbox{%
%     \begin{minipage}{0.925\marginparwidth}
%       \textbf{Our Contributions:} \\
%       \begin{itemize}[noitemsep,nolistsep,leftmargin=*]
%     \item We create a web UI that uses refined visualizations of confusion matrices to capture preferences over pairwise comparisons. 
%     \item The UI implements the binary-search procedure from~\cite{hiranandani2018eliciting} at the back end that make use of the real-time responses over confusion matrices to elicit a linear performance metric in the cancer diagnosis setup.  
%     \item We perform a user study with ten subjects and elicit their linear performance metrics using the proposed web UI. We evaluate the quality of the recovered metric and 
%     % by comparing their responses to the elicited metric's responses over a set of randomly chosen pairwise comparison queries. The study 
%     also include a post-task, \emph{think-aloud}-style interview regarding the utility of the framework.
%     \item Lastly, using the task results and the post-task interviews, we present guidelines regarding practical implementation of the ME framework that can be used for future research in this direction. 
%       \end{itemize}
%     \end{minipage}}\label{sec:sidebar_contributions} }
    
% ME has so far been developed and studied based on the intuition that users can effectively report confusion matrix comparisons. 

Prior literature on ME has proposed elicitation strategies for binary~\cite{hiranandani2018eliciting}, multiclass~\cite{hiranandani2019multiclass}, and multiclass-multigroup~\cite{hiranandani2020fair, hiranandani2022quadratic} classification settings, which assume the presence of an oracle that provides relative preference feedback over confusion matrices. However, to our knowledge, there are no reported implementations testing the ME framework and the assumption that users can effectively report confusion matrix comparisons. 
In this paper, we bring theory closer to practice by providing a first ever practical implementation of the ME framework and its evaluation.  Our contributions are summarized as follows:

% \begin{marginfigure}[0.1cm]
% % \centering
% % \begin{minipage}{\marginparwidth}
% % \begin{figure}
% %   \includegraphics[scale=0.2]{plots/Presentation4.jpeg}
% %  \end{figure}
% % \end{minipage}
% % \begin{minipage}{\marginparwidth}
% \fbox{%
%   \begin{minipage}{0.95\marginparwidth}
    \begin{itemize}[noitemsep,nolistsep,leftmargin=*]
    \item We propose a visualization for pairwise comparison of confusion matrices that adapts the visualization of individual confusion matrices from Shen et al.~\cite{shen2020designing}. 
    % \item We create a web UI that uses refined visualizations of confusion matrices to capture preferences over pairwise comparisons. 
    \item We then integrate the visualization within a web User Interface (UI)\footnote{The UI is available at \href{http://safinahali.com/elicitation-graphs-static/}{http://safinahali.com/elicitation-graphs-static/}}
    % \footnote{The UI is available at http://safinahali.com/elicitation-graphs-static/} 
    that asks for relative preference feedback over confusion matrices. Furthermore, the UI implements Algorithm 1 from Hiranandani et al.~\cite{hiranandani2018eliciting} and uses real-time responses from the subjects to elicit their performance metrics.
    \item Using the proposed UI, we perform a user study with ten subjects and elicit their linear performance metrics for a cost-sensitive binary classification setting. We evaluate the quality of the recovered metrics and 
    % by comparing their responses to the elicited metric's responses over a set of randomly chosen pairwise comparison queries. The study 
    also conduct post-task, \emph{think-aloud}-style interviews with the subjects. 
    % regarding the utility of the framework.
    \item Lastly, we present guidelines for practical implementation of the ME framework for future research.
    % in this direction. 
      \end{itemize}
\section{Problem Setup and Background}
\label{sec:background}
% \vskip -0.2cm

% \sk{I wonder if we want a more intuitive description of this setting.}
Let $X \in \Xcal$ and $Y \in \{0, 1\}$ represent the input and output random variables respectively (0 = negative class, 1 = positive class), and let
% We assume a dataset of size $n$, $\{(x_i, y_i)\}_{i=1}^n$, generated \emph{iid} from a data generating distribution $ \Pmbb \overset{\text{iid}}{\sim} (X, Y)$. 
% the existence of a data generating distribution $(X, Y) \overset{\text{iid}}{\sim} \Pmbb$, which has generated dataset of size $n$, denoted by $\{(x_i, y_i)\}_{i=1}^n$. 
% Let $f_X$ be the marginal distribution for $\Xcal$. 
$\pi = \Pmbb(Y = 1)$ denote the probability of the positive class. 
% Note that the earlier term is a function of the input $x$; whereas, the latter is a constant. 
We denote a classifier by $h$. 
% and let $\Hcal = \{h : \Xcal \rightarrow [0, 1]\}$ be the set of all classifiers. 
A confusion matrix for a classifier $h$ comprises true positives $(TP(h):=\Pmbb(Y = 1, h = 1))$, false positives $(FP(h):=\Pmbb(Y = 0, h = 1))$, false negatives $(FN(h):=\Pmbb(Y = 1, h = 0))$, and true negatives $(TN(h):=\Pmbb(Y = 0, h = 0))$. 
% and is given by:
% \bequation
% 	C_{11} &= TP(h, \Pmbb) = \Pmbb(Y = 1, h = 1), \nonumber \\
% 	C_{01} &= FP(h, \Pmbb) = \Pmbb(Y = 0, h = 1), \nonumber \\
% 	C_{10} &= FN(h, \Pmbb) = \Pmbb(Y = 1, h = 0), \nonumber \\ 
% 	C_{00} &= TN(h, \Pmbb) = \Pmbb(Y = 0, h = 0). 
% 	\label{eq:components}
% \eequation
% Clearly,  $\sum_{i, j} C_{ij} = 1$. We denote the set of all confusion matrices by $\Ccal = \{C(h, \Pmbb)  : h \in \Hcal\}$. 
% Under the population law $\Pmbb$, 
The components of the confusion matrix can be decomposed as:
$FN(h) = \pi - TP(h)$ and $FP(h) = 1 - \pi - TN(h)$, which 
% This decomposition 
reduces the four dimensional space to two dimensional space, and thus we interchangeably refer to the confusion matrix as confusion vector and denote it by $\cmbf(h) = (TP(h), TN(h))$. 
% We will suppress the dependence on $h$ if it is clear from the context.

% \vspace{-0.2cm}
% We will subsume the notation $h$ if it is implicit from the context.  
% Therefore, the set of confusion matrices can be defined as $\Ccal = \{(TP(h, \Pmbb), TN(h, \Pmbb)) : h \in \Hcal\}$. For clarity, we will suppress the dependence on $\Pmbb$ in our notation. In addition, we will subsume the notation $h$ if it is implicit from the context and denote the confusion matrix by $C = (TP, TN)$. We represent the boundary of the set $\Ccal$ by $\partial\Ccal$. 

In this work, we seek to elicit a \emph{linear} performance metric, which for some weights $\ambf$ is defined as:
% \vspace{-0.2cm}
% \begin{equation}
    $\phi((TP(h), TN(h))) = a_0TN(h) + a_1 TP(h).$ 
%     \label{eq:metric}
% \end{equation}
% \vskip -0.2cm
Since the metrics are scale invariant~\cite{narasimhan2014statistical}, we assume $\Vert \ambf \Vert_1=1$. Hence, the linear metrics can be defined using just one parameter as follows:

\begin{equation}
    \phi( (TP(h), TN(h)) ) \coloneqq a_0 TN(h) + (1-a_0) TP(h), \quad \text{where} \quad a_0 \in [0,1].
    \label{eq:metric}
\end{equation}
One example of linear metrics is weighted-accuracy~\cite{koyejo2014consistent}. Specifically, in this work, we want to elicit the weight parameter $a_0$ using the pairwise comparisons of the form $\1[\cmbf(h_1) \succ \cmbf(h_2)]$.

\subsection{Warm-up on Linear Performance Metric Elicitation Algorithm}
\label{ssec:lpmewarmup}

The ME procedure from Hiranandani et al.~\cite{hiranandani2018eliciting} requires a pre-trained estimate of the conditional class probability, i.e., $\hat \eta(x) = \hat{\Pmbb}(Y = 1 | X = x)$ (e.g., logistic regression model estimated using train data). The classifier that optimizes the linear metric in~\eqref{eq:metric} is given by thresholding $\hat \eta(x)$ as follows~\cite{hiranandani2018eliciting}:
\begin{equation}
h_{a_0}[x] = \1[\hat \eta(x) \geq a_0].
\label{eq:thresholdclassifier}
\end{equation}
The above classifier predicts $1$ when for the input $x$, $\hat \eta(x) \geq a_0$ and predicts 0 otherwise. There are two key observations made by Hiranandani et al.~\cite{hiranandani2018eliciting}. First, since there is a one to one correspondence between a linear performance metric and a threshold, eliciting the oracle's performance metric is equivalent to finding the threshold 
% optimal threshold for classifiers of the type in~\eqref{eq:thresholdclassifier} 
that achieves the best performance according to the oracle. Second, under the pairwise comparison setting, the required threshold can be efficiently found via binary search  due to Proposition 2 of~\cite{hiranandani2018eliciting}. 
% In summary, we have a unimodal graph, which achieves the maximum value at the oracle's optimal threshold $a_0$. Thus, binary search can be applied. 
This led Hiranandani et al.~\cite{hiranandani2018eliciting} to create a binary-search type algorithm (Algorithm 1 in~\cite{hiranandani2018eliciting}) that only uses pairwise comparisons from the oracle to elicit the optimal threshold, and hence, the oracle's linear performance metric.
% The procedure is shown in~Algorithm\ref{alg:slme}. 
The algorithm requires the binary search stopping parameter $\epsilon$ and an oracle with linear metric~\eqref{eq:metric}. We provide the details of the algorithm in Appendix A for completeness. 
\vspace{-0.1cm}
\section{Methodology}
\label{sec:method}
\vskip -0.1cm

\subsection{Choice of Task and Dataset}
\label{ssec:dataset}

% \begin{figure*}
%     \centering
%     \includegraphics[scale=0.45]{plots/vis-prior.PNG}
%     \vspace{-0.3cm}
%     \caption{Flow-chart and bar-chart based visualizations for (binary classification) confusion matrices  
%     % in the recidivism prediction task
%     from Shen et al.~\cite{shen2020designing}.}
%     \label{pme-fig:vis-prior}
%     \vskip -0.5cm
% \end{figure*}

Our choice of task is \emph{cancer diagnosis}~\cite{yang2014multiclass} for which we use the Breast Cancer Wisconsin dataset~\cite{dubey2016analysis}. The dataset has two classes, 
% been extensively used in the literature for binary classification, 
where the classes $1$ and $0$ denote \emph{malignant} and \emph{benign} cancer, respectively. There are 699 samples in total, wherein each sample has 9 features. The task for any classifier is to take the 9 features of a patient as input and predict whether or not the patient has cancer. We divide this data into two equally sized parts -- the train and the test data. Using the train data, we fit a logistic regression model to obtain an estimate of the conditional class probability, i.e., $\hat\eta(x) = \hat\Pmbb(Y=1 | X)$. We then create thresholding classifiers of the type: $h_\tau(x) = \1[\hat\eta(x)\geq \tau],$
% \vspace{-0.2cm}
% \begin{equation}
%     h_\tau(x) = \1[\hat\eta(x)\geq \tau],
% \label{eq:classifiers}
% \end{equation}
% \vskip -0.4cm
where we vary $\tau$ from $0$ to $1$ in steps of $1e^{-4}$. The confusion vectors for these thresholded classifiers computed on test data form our query set. That is, our queries are of the form $\1[c(h_{\tau_1}) \succ c(h_{\tau_2})]$ for two thresholds $\tau_1$ and $\tau_2$. We use oracle responses on such queries in the binary-search algorithm to find the optimal threshold $\tau^*$, according to the oracle. 

% as required for running the elicitation procedure from~\cite{hiranandani2018eliciting}. See Figure~\ref{pme-fig:confusions} in Appendix~\ref{append:ssec:dataset}. 

% We then create a sphere using Algorithm~\ref{alg:sphere} inside the space of confusion matrices computed on the test data . 

\subsection{Visualization of Pairwise Comparison}
\label{pme-ssec:vis}
\vskip -0.1cm

\begin{figure*}
% \vspace{-0.5cm}
% \hspace{-0.2cm}
\centering
\includegraphics[scale=0.375]{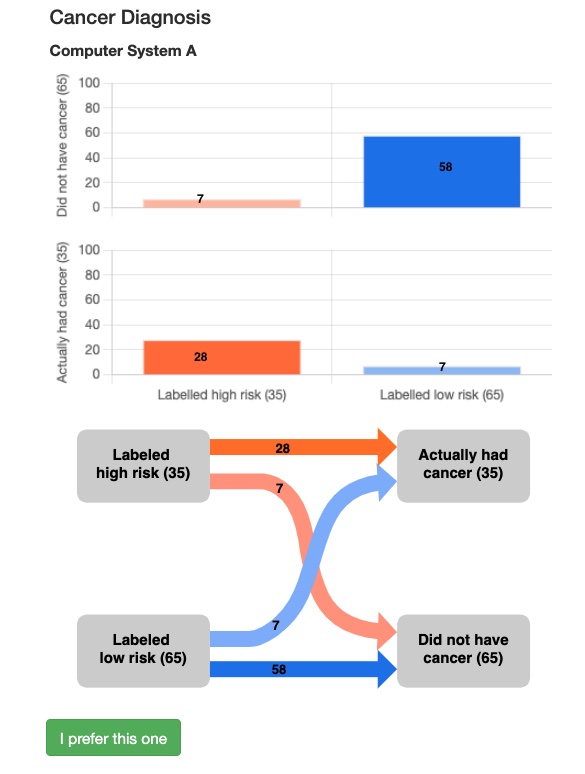}
    \hspace{-0.38cm}
    \includegraphics[scale=0.375]{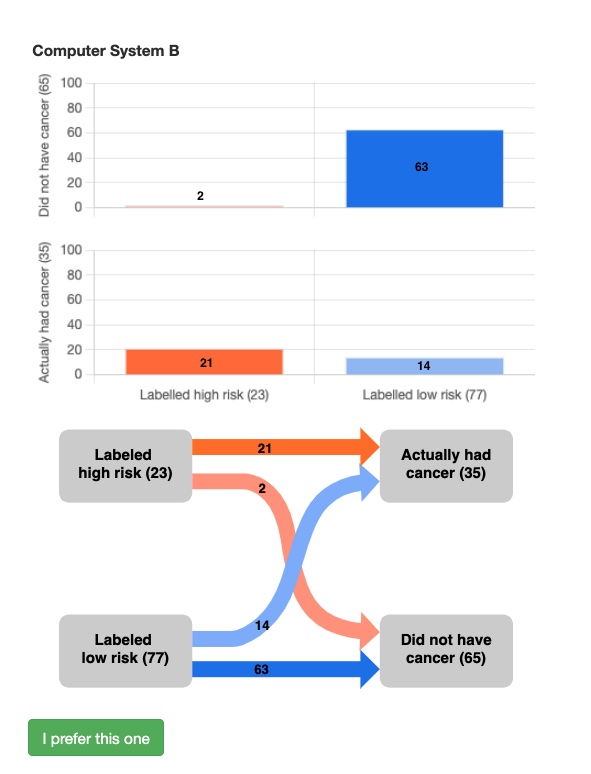}
    \vspace{-0.3cm}
    \caption{Proposed visualization for the pairwise comparison of confusion matrices. The web UI asks for a few comparison queries of such type to the subjects and uses their real-time responses to elicit their  metrics.}
    \vspace{-0.4cm}
  \label{pme-fig:me}
% \end{wrapfigure}
\end{figure*}

% In modern times, ensuring effective public understanding of algorithmic decisions, especially, machine learning models has become an imperative task. Shen et al.~\cite{shen2020designing} provide a concrete step towards this goal by redesigning confusion matrices to support non-experts in understanding the performance of machine learning models. 
Recently, Shen et al.~\cite{shen2020designing} proposed visualizations for confusion matrices to support non-experts in understanding the performance of machine learning models. 
The authors provide four types of visualizations of confusion matrices.
% with distinct information. 
% The visualizations were created over multiple iterative user-studies. 
To find the best visualization out of those four, Shen et al.~\cite{shen2020designing} also conducted a user study. 
% We consider the top two performing visualizations from the user study conducted by Shen et al.~\cite{shen2020designing}. 
% One is the \emph{flow-chart}, which helps in understanding the direction of the confusion entries, and the other is the \emph{bar chart}, which helps in understanding the quantities involved. 
We adapt these visualizations for the context of ME and make the following changes for our pairwise comparison setting (see Figure~\ref{pme-fig:me} for an illustration):
% \vspace{-0.25cm}
% However, in light of our preliminary discussions with Human-Computer Interaction (HCI) and machine learning researchers, we make and recommend the following changes in the visualization for pairwise comparison purposes in the metric elicitation framework.

% \begin{figure}[t]
%     \centering
%     \includegraphics[scale=0.75]{plots/vis-our.PNG}
%     \caption{Our modified visualization of a confusion matrix for a cancer diagnosis task. Modification is  from the perspective of obtaining better pairwise preferences.}
%     \label{pme-fig:vis-our}
% \end{figure}

\begin{enumerate}[noitemsep, nolistsep, leftmargin=*]
    \item Since multiple visualizations of the same information aid in better understanding~\cite{mazza2009introduction}, we choose to use the top two performing visualizations from~\cite{shen2020designing} together to depict a confusion matrix. One is the \emph{flow-chart}, which helps in understanding the direction of the confusion entries, and the other is the \emph{bar chart}, which helps in understanding the quantities involved. 
    \item A query is depicted by presenting two confusion matrices side- by-side, with a button ``I prefer this one'' below each of them asking for the user preference.
    \item We transform the data statistics to denote out-of-100 samples (percentages) for easy understanding.  
    \item Additionally, we show the total number of positive and negative samples along with total number of positive and negative predictions in the visualizations. 
    % are very helpful in comparing two confusion matrices. Therefore, we add the total numbers in the flow-chart boxes and on axes in the bar-charts. 
    % \item We add a zoom-in feature to both the charts. 
    % for better understanding.
    % \item Although, in this user study, we have not changed the direction in the flow-chart, in our discussions with HCI and machine learning researchers, we also noted that the current direction is perhaps more important for the recidivism task (that is because there is time component involved with it) but can be changed for the cancer diagnosis task. This allows one to have constants (i.e., total positive and negative labels) in the left column and the varying component (i.e., total positive and negative predictions) on the right column making the comparison easier. Moreover, this change ensures that the bar-chart and the flow-chart represent similar information. We plan to implement this change and record its impact in our future user studies. 
\end{enumerate}
\vskip -0.25cm
% Our modified visualization incorporating the points above for a confusion matrix in the context of cancer diagnosis is shown in Figure~\ref{pme-fig:me}. 
% A sample of a pairwise comparison query with modified visualizations is shown in Figure~\ref{pme-fig:me}. 
% We next discuss the web user interface. 

\vspace{-0.1cm}
\section{User Interface for Metric Elicitation}
\label{sec:ui}
\vskip -0.1cm

% \begin{wrapfigure}{r}{0.75\textwidth}

% We discuss our proposed web User Interface (UI) in detail and discuss our rationale behind its several components. We also provide images of the UI in Appendix~\ref{tofill}.

Our proposed UI starts with a questionnaire asking about demographic information including  
% like age, gender, race, highest level of school, and 
expertise in machine learning and healthcare.
% as shown in Figure~\ref{pme-fig:index}. 
The UI then broadly has three parts outlined as follows: 
% in the following sub-sections.

% \begin{enumerate}[noitemsep, nolistsep, leftmargin=*]
\vspace{-0.1cm}
    \textbf{Phase-I: Understanding the Visualizations.} The UI first describes the task of cancer diagnosis and how classifiers can be inaccurate in their predictions. Then, it shows a few confusion matrices and asks questions related to \emph{comprehension, comparison, and simulation} (see Table 1 in \cite{shen2020designing}) in the context of cancer diagnosis. This phase familiarize the subjects with the visualizations.
    % and the components associated with the correct and incorrect predictions.
    
    \vspace{-0.1cm}
     \textbf{Phase-II: Eliciting Linear Metrics.} This phase ask subjects for pairwise preferences over confusion matrices and implement the binary-search based procedure, i.e., Algorithm 1 in \cite{hiranandani2018eliciting}. In this work, we fix the stopping parameter $\epsilon$ to $0.05$. 
    %  (detailed in Algorithm~\ref{alg:slme} in Appendix~\ref{append:sec:slme}). 
    %  The confusion matrices are taken from the query space discussed in Section~\ref{ssec:dataset}. 
    %  thus, 
    %  The subjects make a choice reflecting on the trade-off between true positives and true negatives. 
     The UI takes in real-time responses of the subjects reflecting on their trade-off between True Positives and True Negatives, generates next set of queries based on the current responses, and converge to a linear performance metric at the back end. 
    %  We save this (linear) performance metric for each subject. 
    % We stop the binary-search when the search interval becomes less than or equal to 0.05 ($\epsilon$ in line 3 of Algorithm 1 of~\cite{hiranandani2018eliciting}). Moreover, in practice, we do not need to ask four queries per round of binary search; instead, we can reduce the search interval into half by just using at most three pairwise queries in each round. 
    
    \vspace{-0.1cm}
    \textbf{Phase-III: Random Comparisons for Evaluation.} For evaluation purposes, the third phase asks for pairwise preferences on a random set of queries right after the binary search algorithm  has converged, and we have elicited the metric. The subjects are not made aware of this information and are shown queries in continuation to the previous phase.

\vspace{-0.1cm}
\section{User Study and Results}
\label{sec:userstudy}
\vskip -0.1cm

We recruited ten subjects for the study whose demographics are shown in Table~\ref{pme-tab:questionnaire}.  
% (Appendix~\ref{append:sec:extension}). 
The study was approved by IRB at University of Illinois. 
% All the subjects had experience either in biomdical research or machine learning. 
The study was conducted over a video call, where the participants were asked to share the screen after they had filled the questionnaire. 
% The rest of the responses regarding the confusion matrices were over screen share and were logged in the UI. 
% After the elicitation task, 
% the web UI showed a `thank you' page and asked the subjects to close the web browser and screen share. 
After the elicitation task, we asked \emph{think-aloud} style interview questions, which are shown in Table~\ref{pme-tab:posttask}. 

\subsection{Quantitative Results and Findings}
\label{pme-ssec:quant}

% We show the elicited linear metrics of the subjects using the second phase of the UI in Table~\ref{} 
% next discuss the metrics that were elicited for the ten subjects using our web UI, which runs the binary-search based procedure from Algorithm 1 of~\cite{hiranandani2018eliciting} at the back end. 
Once the UI elicits the subject's linear performance metric in Phase-II, it asks for fifteen random queries for evaluation in Phase-III. We measure the fraction of times our elicited metric's preferences matches with the subject's preferences over the fifteen queries in Phase-III, i.e.,\vspace{-0.1cm}
\begin{align*}
\Mcal := \frac{\sum_{i=1}^{15}  \1[\text{subject's preference }  == \text{metric's preference for query } i]}{15}\times 100.
    \label{pme-eq:fraction}
\end{align*}
\vskip -0.3cm

We show the elicited metric for the ten subjects and the $\Mcal$ values in Table~\ref{pme-tab:metrics}. We see for nine out of ten subjects that more than 85\% of the times our elicited metric's preferences matches with the subject's preferences on the fifteen evaluation queries. The absolute $\Mcal$ measure suggest that our approach is effective; however, without comparing to a baseline, it is not yet known how effective it really is. 
% still a missing piece in this study because of the lack of a baseline. 
In future, we plan to develop a baseline for comparison on the ME task using the measure $\Mcal$.

\subsection{Qualitative Feedback and Guidelines for the Future}
\label{pme-ssec:qual}

We now summarize feedback observed during the user study and the findings from the interviews. We also formulate guidelines for future research on practical ME, which are shown in Table~\ref{pme-tab:guidance}. 
% For subjects' excerpts and detailed qualitative discussion, see Appendix~\ref{append:ssec:qual}. 

\textbf{Observations during study sessions:} We noted that subjects were not comfortable in answering the \emph{simulation}-based questions in Phase-I of the UI (Section~\ref{sec:ui}). A possible reason could be that the direction of the flow-chart is opposite to the conditioning of probability that is asked in those questions (\textbf{G1}, Table\ref{pme-tab:guidance}). 
% Bar-chart allows them to answer this question easily; however, we find that by this point in the UI, the subject becomes more comfortable with using the flow-chart. 
% Some users when asked in the post-interview session also mentioned that this could help them better in the pairwise comparison, too. 
While comparing confusion matrices, we also noted that after a few rounds, the subjects tend to look at only the flow-charts for comparison (\textbf{G2}, Table\ref{pme-tab:guidance}). 

\begin{table}[t]
\parbox{.35\linewidth}{
\vspace{-0.05cm}
\centering
    \caption{Subjects' demographics: The values in parenthesis show the number of subjects. ML and Healthcare stand for ML and healthcare knowledge, respectively.}
    \vspace{0.175cm}
    \scriptsize{
    \begin{tabular}{|@{}c@{}|@{}c@{}|@{}c@{}|@{}c@{}|}
    \hline
         \textbf{Age} & \textbf{Education} & \textbf{ML} & \textbf{HC} \\
         25(2) & Grad. Col.(4) & None(5) & None(5) \\ 
         26(3) & Masters(3) & Begin.(3) & Some(2) \\
         28(5) & Ph.D.(3) &  Intermed.(2) & No resp.(3) \\
        %  & 25 (2) & 26 (3) & 28 (5) \\
        %  \textbf{Education Level} & Graduate College (4) & Master's (3) & Doctorate (3) \\
        %  \textbf{ML Expertise} & None (5) & Beginner (3) & Intermediate (2) \\
        %  \textbf{Healthcare Knowledge} & None (5) & Some (2) & No response (3)\\
    \hline
    \end{tabular}
    \label{pme-tab:questionnaire}
    }
    }
% \hfill 
\hspace{.1cm}
\parbox{.6\linewidth}{
\scriptsize{
\centering
\caption{Post-task interview questions.}
    \begin{tabular}{|p{0.035\linewidth}|p{0.91\linewidth}|}
    \hline
         \textbf{Q1} &  What do you think is worse: (a) Large number of patients that actually have cancer but are labelled as low risk, or 
         %by a computer system, or 
(b) Large number of patients that do not have cancer but are labelled as high risk. 
% by a computer system.
\\
         \textbf{Q2} & Could you quantify how much worse the chosen option is in comparison to the other? Why or why not? Could you quantify this? i.e, 10x worse for me
 \\
         \textbf{Q3} & For the questions presented in this task, how did you decide which system you would prefer your doctor to use?
 \\
         \textbf{Q4} & What was difficult about making these choices?
 \\
         \textbf{Q5} & What additional information would have helped you to make these choices?
 \\
         \textbf{Q6} & Do you have any feedback for us on your experience today? 
\\
    \hline
    \end{tabular}
    \label{pme-tab:posttask}
    }
    }
\end{table}

\begin{table}[t]
\parbox{.35\linewidth}
{
\vspace{-0.7cm}
\centering
\caption{Elicited metrics for the subjects along with $\Mcal$.
    % fraction of times (in \%) the elicited metric's preferences matches with the subject's preferences over the fifteen evaluation queries.
    }
    \vspace{-0.1cm}
\scriptsize{
\centering
    \begin{tabular}{|c|c|c|}
    \hline
    \textbf{S} & \textbf{Metrics} & $\Mcal$ \\
    \hline
         S1 & 0.125 \text{TN} + 0.875 \text{TP}  & 87\\
         S2 & 0.141 \text{TN} + 0.859 \text{TP}  & 100\\
         S3 & 0.125 \text{TN} + 0.875 \text{TP}  & 93\\
         S4 & 0.141 \text{TN} + 0.859 \text{TP}  & 100\\
         S5 & 0.328 \text{TN} + 0.672 \text{TP}  & 73\\
         S6 & 0.031 \text{TN} + 0.969 \text{TP}  & 87\\
         S7 & 0.031 \text{TN} + 0.969 \text{TP}  & 100\\
         S8 & 0.359 \text{TN} + 0.641 \text{TP}  & 87\\
         S9 & 0.125 \text{TN} + 0.875 \text{TP}  & 93\\
         S10 & 0.141 \text{TN} + 0.859 \text{TP}  & 87\\
    \hline
    \end{tabular}
    % \vspace{-0.25cm}
    \label{pme-tab:metrics}
    \vskip -0.6cm
    }
}
% \hfill 
\hspace{0.1cm}
\parbox{.6\linewidth}
{
% \vspace{-1cm}
\caption{Guidelines for ME.}
    \scriptsize{
    \begin{tabular}{|@{}c@{}|p{0.95\linewidth}|}
    \hline
        %  \textbf{G1} &  Whenever possible, smoothen the query space so to run the binary-search based algorithms with reduced finite sample errors.\\
         \textbf{G1} &  
        The direction in the flow-chart 
        %  based visualization of the confusion matrix 
         can be swapped with total number of labels shown in the left column and total predictions on the right. \\
        %  in a query.\\
         \textbf{G2} & Showing only flow-chart for pairwise comparisons is better than showing flow-chart and bar-chart together. \\
        %  \textbf{G3} & %  Depending on the search tolerance of the binary-search, 
        %  Show probabilities in the confusion matrix as out-of-$n$ samples, where bigger the $n$, the better it is to differentiate between confusion matrices. \\
        %  One may also just show, the false positives and false negatives to further reduce the information load.\\
        %  \textbf{G5} & Measure time to respond for each query. Spending more time on queries that comprise close confusion matrices lead credence to the noise model in Definition~\ref{me-def:noise}.\\
        %  \textbf{G4} & The terminology ``labelled as high risk/low risk" can be replaced with ``predicted as high risk/low risk" to avoid confusions regarding ground-truth label.\\
         \textbf{G3} & Corresponding to the post-task interview question 2, one needs to devise a UI so to ask for the intuitive guess for the false negative cost. This would also act as a baseline method for evaluation purposes as discussed in Section~\ref{pme-ssec:quant}. \\
        %  (see Section~\ref{pme-ssec:quant}).\\
         \textbf{G4} & In addition to confusion entries such as true positives, one should show percentages conditioned on the true classes, i.e., true positive rates. \\
        %  This would aid in making comparisons.\\
         \textbf{G5} & Extend the description 
        %  on cancer diagnosis and mention 
         of the associated subjective costs for the incorrect predictions (e.g., financial, emotional, etc.) from the task perspective. \\
        %  \gh{Write Conclusions} \\
        %  or excerpts that cover different aspects of the cost. \\
        %  For example, how much financial burden a false positive prediction would put on a patient, how much emotional burden would it put, what are the possible side-effects of drugs, etc. 
    \hline
    \end{tabular}
    % \vspace{-0.2cm}
    \label{pme-tab:guidance}
    }
}
% \vskip -0.4cm
\end{table}
    
% \begin{table}[t]
% \caption{Elicited linear performance metrics for the ten subjects along with the measure $\Mcal$.
%     % fraction of times (in \%) the elicited metric's preferences matches with the subject's preferences over the fifteen evaluation queries.
%     }
% \scriptsize{
% \centering
%     \begin{tabular}{|c|c|c|}
%     \hline
%     \textbf{S} & \textbf{Metrics} & $\Mcal$ \\
%     \hline
%          S1 & 0.125 \text{TN} + 0.875 \text{TP}  & 87\\
%          S2 & 0.141 \text{TN} + 0.859 \text{TP}  & 100\\
%          S3 & 0.125 \text{TN} + 0.875 \text{TP}  & 93\\
%          S4 & 0.141 \text{TN} + 0.859 \text{TP}  & 100\\
%          S5 & 0.328 \text{TN} + 0.672 \text{TP}  & 73\\
%          S6 & 0.031 \text{TN} + 0.969 \text{TP}  & 87\\
%          S7 & 0.031 \text{TN} + 0.969 \text{TP}  & 100\\
%          S8 & 0.359 \text{TN} + 0.641 \text{TP}  & 87\\
%          S9 & 0.125 \text{TN} + 0.875 \text{TP}  & 93\\
%          S10 & 0.141 \text{TN} + 0.859 \text{TP}  & 87\\
%     \hline
%     \end{tabular}
%     % \vspace{-0.25cm}
%     \label{pme-tab:metrics}
%     \vskip -0.6cm
%     }
% \end{table}

\textbf{Post-task interview sessions:} In response to Q1 (Table~\ref{pme-tab:posttask}), every subject clearly figured out the direction of the costs. 
% and mentioned that (in the words of S1), \emph{``a patient who has cancer but was predicted as low risk is a costlier mistake than a patient who does not have cancer but was predicted as high risk."}
However, none of the subjects could answer Q2 (Table~\ref{pme-tab:posttask}) with full confidence. This further suggests the effectiveness of the ME framework. 
% Often, practitioners make a guess to quantify the asymmetric costs in class-imbalanced learning; however, the guess may be far from innate costs of the practitioner. 
The subjects agreed that it is easier to compare two confusion matrices using the proposed UI than to directly quantify the costs (\textbf{G3}, Table\ref{pme-tab:guidance}). In response to Q5 (Table~\ref{pme-tab:posttask}), subjects mentioned that having quantities such as true positive rate in addition to true positives, would be helpful in making comparisons (\textbf{G4}, Table\ref{pme-tab:guidance}). Lastly, subjects mentioned that it would have been easier to compare if some subjective description of the associated costs for incorrect predictions were provided in the beginning of the study (\textbf{G5}, Table\ref{pme-tab:guidance}).

% \sk{Should there be a conclusion, or "discussion"?}
% \sk{Shoudl there be something about the IRB? Either here or in acknowledgements? Not sure...}
\vspace{-0.2cm}
\section{Conclusion}
\label{sec:conclusion}
\vskip -0.15cm

We created a web user-interface to practically elicit user performance metrics in a binary classification setting. 
% We chose cancer diagnosis as the task domain, because it involves asymmetric costs for false positives and false negatives. 
% We adapted existing visualizations of confusion matrices to capture preferences over pairwise comparisons. 
Via a user study with ten subjects, we demonstrated an implementation of the metric elicitation procedure from~\cite{hiranandani2018eliciting} that makes use of the real-time user responses. 
% over pairwise comparisons of confusion matrices. 
We also proposed an evaluation scheme to judge the quality of the recovered metric. 
% Using the proposed web UI, we then conducted a user study with ten subjects and elicited their linear performance metrics. We also compared the quality of the recovered metric  by comparing their responses to the elicited metric's responses over a set of randomly chosen pairwise comparison queries. The study also included a post-task, \emph{think-aloud}-style interviews regarding the utility of the framework. 
Lastly, using the study findings, we formulated guidelines for practical implementation of the ME framework. 
% In the future, we plan to build upon this study and conduct a comprehensive user study that includes the guidelines presented in this paper with more subjects. 
% We also plan to extend the current web UI to elicit metrics in the multiclass classification setup.

\bibliographystyle{plain}
\bibliography{sample}

\begin{thebibliography}{10}

\bibitem{Dmitriev2016MeasuringM}
Pavel Dmitriev and Xian Wu.
\newblock Measuring metrics.
\newblock In {\em CIKM}, 2016.

\bibitem{dubey2016analysis}
Ashutosh~Kumar Dubey, Umesh Gupta, and Sonal Jain.
\newblock Analysis of k-means clustering approach on the breast cancer
  wisconsin dataset.
\newblock {\em International journal of computer assisted radiology and
  surgery}, 11(11):2033--2047, 2016.

\bibitem{hiranandani2018eliciting}
Gaurush Hiranandani, Shant Boodaghians, Ruta Mehta, and Oluwasanmi Koyejo.
\newblock Performance metric elicitation from pairwise classifier comparisons.
\newblock In {\em The 22nd International Conference on Artificial Intelligence
  and Statistics}, pages 371--379, 2019.

\bibitem{hiranandani2019multiclass}
Gaurush Hiranandani, Shant Boodaghians, Ruta Mehta, and Oluwasanmi~O Koyejo.
\newblock Multiclass performance metric elicitation.
\newblock In {\em Advances in Neural Information Processing Systems}, pages
  9351--9360, 2019.

\bibitem{hiranandani2022quadratic}
Gaurush Hiranandani, Jatin Mathur, Harikrishna Narasimhan, and Oluwasanmi
  Koyejo.
\newblock Quadratic metric elicitation for fairness and beyond.
\newblock In {\em Uncertainty in Artificial Intelligence}, pages 811--821.
  PMLR, 2022.

\bibitem{hiranandani2020fair}
Gaurush Hiranandani, Harikrishna Narasimhan, and Oluwasanmi Koyejo.
\newblock Fair performance metric elicitation.
\newblock In {\em NeurIPS}, 2020.

\bibitem{koyejo2014consistent}
Oluwasanmi~O Koyejo, Nagarajan Natarajan, Pradeep~K Ravikumar, and Inderjit~S
  Dhillon.
\newblock Consistent binary classification with generalized performance
  metrics.
\newblock In {\em NIPS}, pages 2744--2752, 2014.

\bibitem{mazza2009introduction}
Riccardo Mazza.
\newblock {\em Introduction to information visualization}.
\newblock Springer Science \& Business Media, 2009.

\bibitem{narasimhan2014statistical}
Harikrishna Narasimhan, Rohit Vaish, and Shivani Agarwal.
\newblock On the statistical consistency of plug-in classifiers for
  non-decomposable performance measures.
\newblock In {\em Advances in Neural Information Processing Systems}, pages
  1493--1501, 2014.

\bibitem{qian2013active}
Buyue Qian, Xiang Wang, Fei Wang, Hongfei Li, Jieping Ye, and Ian Davidson.
\newblock Active learning from relative queries.
\newblock In {\em IJCAI}, pages 1614--1620, 2013.

\bibitem{shen2020designing}
Hong Shen, Haojian Jin, {\'A}ngel~Alexander Cabrera, Adam Perer, Haiyi Zhu, and
  Jason~I Hong.
\newblock Designing alternative representations of confusion matrices to
  support non-expert public understanding of algorithm performance.
\newblock {\em Proceedings of the ACM on Human-Computer Interaction},
  4(CSCW2):1--22, 2020.

\bibitem{yang2014multiclass}
Sitan Yang and Daniel~Q Naiman.
\newblock Multiclass cancer classification based on gene expression comparison.
\newblock {\em Statistical applications in genetics and molecular biology},
  13(4):477--496, 2014.

\bibitem{zhang2020joint}
Yunfeng Zhang, Rachel Bellamy, and Kush Varshney.
\newblock Joint optimization of ai fairness and utility: A human-centered
  approach.
\newblock In {\em Proceedings of the AAAI/ACM Conference on AI, Ethics, and
  Society}, pages 400--406, 2020.

\end{thebibliography}

\clearpage
\onecolumn
\appendix

% \onecolumn

% \begin{appendices}
\begin{center}
\textbf{\Large Appendix for ``Metric Elicitation; Moving from Theory to Practice''}
\end{center}
~\\

% \section{Linear Performance Metric Elicitation (LPME)}
% \label{append:sec:slme}

\begin{figure}[h]
\begin{minipage}[h]{\textwidth}
  \centering \hspace{-0.5em}
  \begin{minipage}[h]{.52\textwidth}
     \centering
\fbox{\parbox[t]{1\textwidth}{\vspace{0.0cm}\small{\underline{\bf Subroutine \emph{ShrinkInterval}}\normalsize}    \\
\small
\textbf{Input:} Oracle responses for $\Omega(\cmbfbar^{(c)}, \cmbfbar^{(a)})$,\\
$\Omega(\cmbfbar^{(d)}, \cmbfbar^{(c)}),$ $\Omega(\cmbfbar^{(e)}, \cmbfbar^{(d)}), \Omega(\cmbfbar^{(b)},  \cmbfbar^{(e)})$\\
%%%%%%%% binary search part %%%%%%%%%%%%%
\textbf{If} \, ($\cmbfbar^{(a)} \succ \cmbfbar^{(c)}$) Set $\tau^{(b)} = \tau^{(d)}$.\\
\textbf{elseif} \, ($\cmbfbar^{(a)} \prec \cmbfbar^{(c)} \succ \cmbfbar^{(d)}$) Set $\tau^{(b)} = \tau^{(d)}$.\\
\textbf{elseif} \, ($\cmbfbar^{(c)} \prec \cmbfbar^{(d)} \succ \cmbfbar^{(e)}$) Set $\tau^{(a)} = \tau^{(c)}$,  $\tau^{(b)} = \tau^{(e)}$.\\
\textbf{elseif} \, ($\cmbfbar^{(d)} \prec \cmbfbar^{(e)} \succ \cmbfbar^{(b)}$) Set $\tau^{(a)} = \tau^{(d)}$.\\
\textbf{else} 
 Set $\tau^{(a)} = \tau^{(d)}$.\\
% \text{ \ \ \ \ }\textbf{end}\\
%%%%%%% binary search part %%%%%%%%%%%%%
\textbf{Output:} $[\tau^{(a)}, \tau^{(b)}]$.  
\normalsize \vspace{-0.07cm}
}}
  \end{minipage} \hspace{0.3em}
  \begin{minipage}[h]{.45\textwidth}
     \centering
    %  \captionof{algorithm}{LPM Elicitation}
    %  \label{alg:alg2}
    %  \begin{algorithmic}
    %   \addtocounter{algorithm}{1}
% \balgorithm[t]
% \caption{LPM Elicitation}
% \label{alg:linear}
\centering
\includegraphics[scale=0.5]{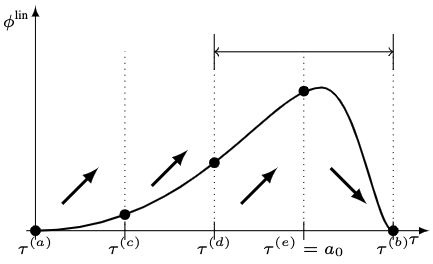}
  \end{minipage}
%   \captionof{figure}{Two algorithms side by side}
\end{minipage}
\caption{(Left): The \emph{ShrinkInterval} subroutine used in line 7 of Algorithm~\ref{alg:slme} (Right): Visual illustration of the subroutine \emph{ShrinkInterval}; \emph{ShrinkInterval} shrinks the current search interval to half based on oracle responses to four queries.}
\vskip -0.4cm
\label{append:fig:shrink1}
\end{figure}

The following procedure is borrowed from~\cite{hiranandani2018eliciting} and described here for completeness. 
\balgorithm[h]
\caption{Linear Performance Metric Elicitation}
\label{alg:slme}
\small
\balgorithmic[1]
\STATE \textbf{Input:}  Binary-search tolerance $\epsilon > 0$, oracle $\Omega(\cdot, \cdot\,;\, \phi^{\text{lin}})$ with metric $\phi^{\text{lin}}$\\ \hfill\\
% \FOR{$i = 1, 2, \cdots q$} 
% \STATE Set $\ambf = \ambf' = (1/\sqrt{q}, \dots, 1/\sqrt{q})$.
% \STATE Set $a'_i = -1/\sqrt{q}$.
% \STATE Compute the optimal $\sbar^{(\ambf)}$ and $\sbar^{(\ambf')}$ over the sphere $\Scal$ using Lemma~\ref{lem:spherebayes}
% \STATE Query $\Omega(\smbfbar^{(\ambf)}, \smbfbar^{(\ambf')} ; \phi^{\text{lin}})$\\
% \ENDFOR
% \COMMENT{These queries reveal the search orthant}\\ \hfill \\
% \STATE Start with coordinate $j=1$.
% \STATE\textbf{Initialize:} $\bm{\theta} = \bm{\theta}^{(1)}$ \hfill \COMMENT{$\bm{\theta}^{(1)}$ is a point in the search orthant.}
% \FOR{$t=1, 2, \cdots, T=4(q-1)$}
% \STATE Set $\bm{\theta}^{(a)} = \bm{\theta}^{(c)}=\bm{\theta}^{(d)}=\bm{\theta}^{(e)}=\bm{\theta}^{(b)} = \bm{\theta}^{(t)}$.\\
% % \IF{$ t\%(q-1)$}
% % \STATE Set $j = t\%(q-1)$
% % \ELSE
% % \STATE $j = q-1$
% % \ENDIF
% % \IF{$j == q - 1$} 
% % \STATE \textbf{Initialize:} $\theta^a_{j} = \pi$, $\theta^b_j = 3\pi/2$.
% % \ELSE
% % \STATE \textbf{Initialize:} $\theta^a_j = \pi/2$, $\theta^b_j = \pi$
% % \ENDIF
\STATE Set $\tau^{(a)} = 0$ and $\tau^{(b)} = 1$
\WHILE{$\abs{\tau^{(b)} - \tau^{(a)}} > \epsilon$}
\STATE Set $\tau^{(c)} = \frac{3 \tau^{(a)} + \tau^{(b)}}{4}$, $\tau^{(d)} = \frac{\tau^{(a)} + \tau^{(b)}}{2}$, and $\tau^{(e)} = \frac{\tau^{(a)} + 3 \tau^{(b)}}{4}$.
\STATE Set $\cmbfbar^{(a)}, \cmbfbar^{(c)}, \cmbfbar^{(d)}, \cmbfbar^{(e)},$ and $\cmbfbar^{(b)}$ to be the confusion matrices corresponding to the thresholds $\tau^{(a)}, \tau^{(c)}$, $\tau^{(d)}$, $\tau^{(e)}$, and $\tau^{(b)}$, respectively.
\STATE Query $\Omega(\cmbfbar^{(c)}, \cmbfbar^{(a)}), \quad \Omega(\cmbfbar^{(d)},  \cmbfbar^{(c)})$, $\quad$ $\Omega(\cmbfbar^{(e)}, \cmbfbar^{(d)}), \quad \Omega(\cmbfbar^{(b)}, \cmbfbar^{(e)})$.
\STATE $[\tau^{(a)}, \tau^{(b)}] \leftarrow$ \emph{ShrinkInterval} (based on user responses)\hfill \COMMENT{see Figure~\ref{append:fig:shrink1}}
\ENDWHILE
\STATE Set $\tau^{(d)} = \frac{1}{2}(\tau^{(a)}+\tau^{(b)})$ \\
\STATE \textbf{Output:} $\tau^{(d)}$
\ealgorithmic
\ealgorithm

The \emph{ShrinkInterval}  subroutine (illustrated in Figure~\ref{append:fig:shrink1}) is binary-search based routine that shrinks the interval $[\tau^{(a)}, \tau^{(b)}]$ by half based on the oracle responses to four queries. 

\end{document}